\documentclass[a4paper, amsfonts, amssymb, amsmath, reprint, showkeys, nofootinbib, twoside]{revtex4-1}
\usepackage[english]{babel}
\usepackage[utf8]{inputenc}
\usepackage[colorinlistoftodos, color=green!40, prependcaption]{todonotes}
\usepackage{amsthm}
\usepackage{mathtools}
\usepackage{physics}
\usepackage{xcolor}
\usepackage{graphicx}
\usepackage[left=23mm,right=13mm,top=35mm,columnsep=15pt]{geometry} 
\usepackage{adjustbox}
\usepackage{placeins}
\usepackage[T1]{fontenc}
\usepackage{lipsum}
\usepackage{csquotes}

\usepackage[pdftex, pdftitle={Article}, pdfauthor={Author}]{hyperref} 
\begin{document}
\title{Fast Analysis of the OpenAI O1-Preview Model in Solving Random K-SAT Problem: Does the LLM Solve the Problem Itself or Call an External SAT Solver?}

\author{Raffaele Marino}
    \email[Correspondence email address: ]{raffaele.marino@unifi.it}
    \affiliation{Department of Physics and Astronomy, University of Florence, Via Giovanni Sansone, 1, 50019 Sesto Fiorentino FI}

\date{\today} 

\begin{abstract}
In this manuscript, I present an analysis on the performance of OpenAI O1-preview model in solving random K-SAT instances for K$\in {2,3,4}$ as a function of $\alpha=M/N$ where $M$ is the number of clauses and $N$ is the number of variables of the satisfiable problem. I show that the model can call an external SAT solver to solve the instances, rather than solving them directly. Despite using external solvers, the model  reports incorrect assignments as output. Moreover, I propose and present an analysis to quantify whether the OpenAI O1-preview model demonstrates a spark of intelligence or merely makes random guesses when outputting an assignment for a Boolean satisfiability problem.

\end{abstract}


\maketitle
\section{Introduction} \label{sec:introduction}

Today, Artificial Intelligence (AI) has become a major topic in both academic and industrial research.  It helps us in predicting and forecasting events, as well as generating text and images with high accuracy \cite{bishop2023deep}. The AI technologies that are widely used today are primarily based on deep learning methods, particularly deep neural networks, which learn to make predictions or forecasts for specific tasks \cite{lecun2015deep}. In the domain of text generation, transformers \cite{10.5555/3295222.3295349} are the predominant architectures. These models utilize an attention mechanism to capture complex statistical structures within natural language \cite{bishop2023deep}. Many AI applications have been proposed \cite{7780459, 10.5555/3495724.3496298, Han2017, marino2023solving}, with ChatGPT \cite{bubeck2023sparks, RAY2023121} emerging as one of the most renowned examples globally. This model has shown remarkable accuracy in generating text that can be comparable to that of a highly educated person. However, when it comes to understanding and solving complex mathematical problems, as other transformers,  it has historically been less effective \cite{10.5555/3666122.3667718, 10.5555/3666122.3669203, WANG2023930}.

Recently\footnote{$12^{\text{th}}$ of September 2024}, OpenAI released a new model (O1-preview) that reportedly achieves text generation comparable to that of a Ph.D.-educated person and demonstrates high performance in solving mathematical problems \cite{O1model, Lingo2024}.
For instance, consider the following problem: "There is a plate of candies on a table in a room with three people: Alice, Bob, and Jenny. Bob does not like candies, while Alice and Jenny do. Who could eat the candies?" The correct answer is Alice or Jenny. This simple problem can be formulated as a satisfiability problem \cite{10.1145/2696081.2696091}, where Alice, Bob, and Jenny are represented by three Boolean variables \(x_1, x_2, x_3 \in \{0,1\}\) respectively, while the given conditions can be expressed in a logical clause of the form \((x_1 \lor \neg x_2 \lor x_3)\), where $\lor$ is the logical operator OR. The solution is determined by finding an assignment of these variables that satisfies all the given conditions. When the number of clauses increases, the problem becomes more challenging. In the so-called thermodynamic limit in physics \cite{Marino_2024}, where \(M \to \infty\), \(N \to \infty\), and \(\lim_{{M, N \to \infty}} \alpha < \infty\), the problem undergoes a phase transition at a critical value \(\alpha_{s}\) \cite{marino2016backtracking}. This phase transition indicates that above $\alpha_{s}$ no satisfiable assignment exists—meaning a configuration of Boolean variables that simultaneously satisfies all clauses cannot be found.

A simple problem like the one described above can be easily solved by the OpenAI O1-preview model. Encouraged by this result, I decided to test the performance of OpenAI O1-preview model on simple random K-SAT instances with \(K \in \{2, 3, 4\}\) as a function of \(\alpha\), near the satisfiability threshold \(\alpha_{s}\). The goal was to solve different instances for various values of \(\alpha\) and \(N\) in the spirit of statistical physics \cite{doi:10.1073/pnas.0703685104} for identifying the numerical algorithmic threshold of the model.

However, as the analysis progressed, I observed that as the complexity of an instance increased, the model, rather than solving the problem itself, was calling an external SAT solver (an example is presented in Appendix: SAT solver calling). This rendered the analysis ineffective  and raises significant concerns about the evaluation and verification of the scientific robustness of the OpenAI O1-preview model, particularly given that it is not open-source. 

Without access to the underlying model, it is impossible to determine whether the OpenAI O1-preview model has genuinely learned the functionality and capabilities of a SAT solver or is merely outsourcing the problem to an external tool.  In both scenarios, the model is an AI system; however, in the first case, when the Large Language Model (LLM) can solve a random K-SAT problem on its own, the analysis can be conducted with scientific validity because the system functions as a single, unified entity attempting to solve a complex problem.

In the second case, where the LLM delegates the problem to an external SAT solver, the analysis lacks scientific validity because the performance results are derived from a SAT solver, rather than the LLM itself. In such a case, the AI system results from merging a Machine Learning system and an AI Expert System.

This Letter aims to raise awareness within the scientific community about the limitations of using non-open-source models. By conducting an analysis following a fixed protocol (see Appendix: Protocol), I present the results obtained from the OpenAI O1-preview model and compare them with those from a well known solver like Pycosat\cite{biere2008picosat}. Additionally, I show that the model under study can indeed call an external SAT solver. Moreover, I propose and present an analysis to quantify whether the OpenAI O1-preview model demonstrates a spark of intelligence or merely makes random guesses when outputting an assignment for a Boolean satisfiability problem.

The analysis is intentionally kept as simple as possible to ensure reproducibility. We also release the benchmark dataset used in this study, along with all the solutions obtained and analyzed \footnote{\href{https://github.com/RaffaeleMarino/AnalysisOpenAIO1ModelKSAT}{github.com/RaffaeleMarino/AnalysisOpenAIO1ModelKSAT}} \footnote{All results from the OpenAI O1-preview model were obtained on the $14^{\text{th}}$ and $19^{\text{th}}$ of September, 2024, using three different "Plus" accounts at a total cost of $60\$$ without taxes.}. 

The remaining sections of this paper are organized as follows.  Section \ref{sec:KSAT} recalls briefly the random K-SAT problem and the state-of-the-art of SAT solvers. Section \ref{sec:results} details the analysis performed using the model on random K-SAT instances, while in Section \ref{sec:conclusions}, I conclude the paper discussing and summarizing the obtained results.

\section{K-SAT} \label{sec:KSAT}

The SAT problem, in its most fundamental form, is defined by a set of $N$ Boolean variables. Each Boolean variable is binary, taking one of two values: $\{0,1\}$. Adding complexity to the SAT problem are $M$ constraints, each one forming a clause. A clause is a disjunction (an OR operation $\lor$) of literals, with each literal being either a Boolean variable or its negation. So, if I denote a Boolean variable as $x_i$, then both $x_i$ and $\neg x_i$ qualify as literals. An instance of a SAT problem is a conjunction (an AND operation $\land$) of clause. The task is to find a satisfiable assignment for the boolean variables that makes all clauses true. Given the dichotomous nature of Boolean variables, each clause could have $2^k$ assignments (not all of them satisfiable) where $k$ is the number of literals in the clause.  The broad SAT problem branches into specific subsets. In random K-SAT, each clause is restricted to exactly K literals. The random 2-SAT problem is part of class P, solvable in polynomial time, while random K-SAT, with K$\geq 3$ is NP-Complete, indicating a higher complexity. 

The random K-SAT problem is viewed by statistical physicists as a problem exhibiting the so-called frustration property \cite{ANDERSON1978291}, which is of great interest to physics of spin glasses \cite{VINCENT2024371}. As stated in the introduction, the random K-SAT problem  presents in the thermodynamic limit a phase transitions at $\alpha_s$ that signifies the partitioning of the space of solutions for the random K-SAT problem into two distinctive regions. On one side $\alpha \in [0, \alpha_{s}]$, the problems has feasible solutions, while on the other $\alpha \in (\alpha_{s}, \infty)$, no assignment exists to satisfy all clauses of a given instance. Mathematical studies have confirmed the existence of these phase transitions for various values of K. For instance, the random 2-SAT problem shows a critical threshold value at $\alpha_{s}=1$ \cite{mezard2009information}, while the recent discovery for the random 3-SAT problem places its threshold value at $\alpha_{s}=4.267$ \cite{10.1145/2746539.2746619} and for the random 4-SAT problem the threshold is $\alpha_{s}=9.931$ \cite{Montanari_2008}. Remarkably, this behavior of the probability of finding a satisfiable assignment remains consistent across all different values of K \cite{10.1145/2746539.2746619}.

Many algorithms have been developed for solving the random K-SAT problem. They can be exact, with an exponential complexity (or for random 2-SAT with a polynomial complexity), heuristics, like Pycosat \cite{biere2008picosat, 10.5555/1550723} or Survey Propagation and its variants \cite{parisi2002, marino2016backtracking}, or learning algorithms \cite{Marino_2021} based also on Graph Neural Networks\cite{ZHOU202057}. In this manuscript, for making a comparison with a well established method I use Pycosat \cite{biere2008picosat}.

\section{Results} \label{sec:results}

\begin{figure*}[ht]
    \centering
    \includegraphics[width=0.3\textwidth]{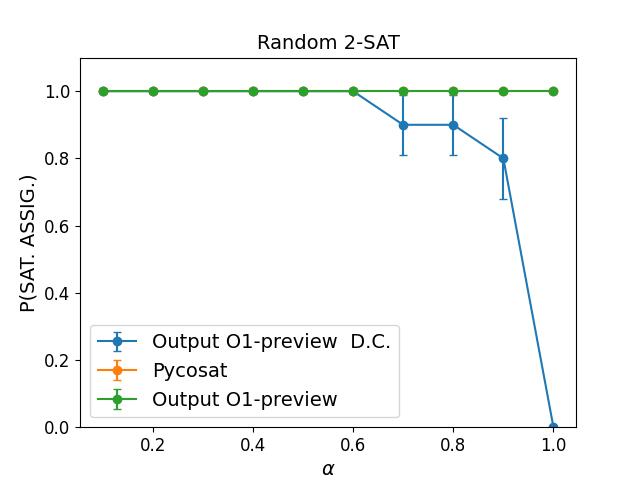} 
    \includegraphics[width=0.3\textwidth]{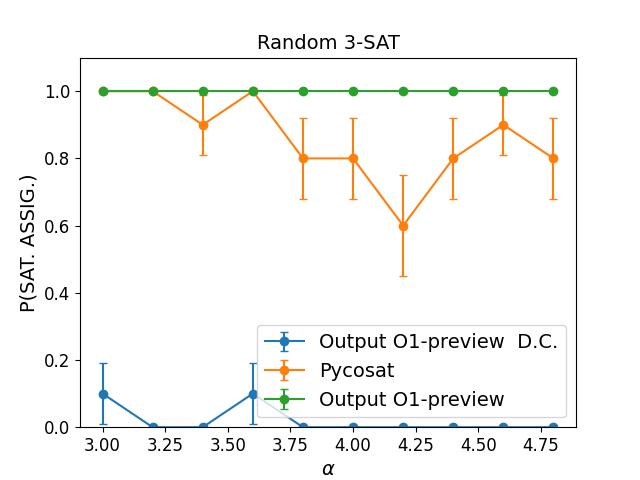} 
    \includegraphics[width=0.3\textwidth]{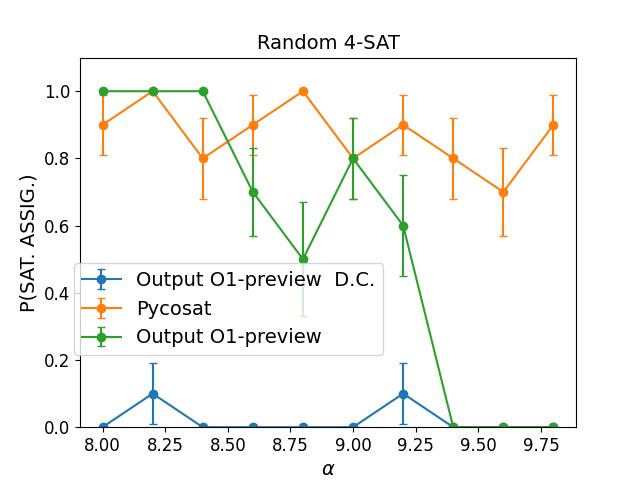} 
    \caption{Fraction of satisfiable assignments, namely P(SAT. ASSIG.), as a function of $\alpha$. Each point is an average over 10 samples and error bars are standard errors. Green points identify the fraction of satisfiable assignments returned and checked by OpenAI O1-preview model, blue points identify the fraction of satisfiable assignments returned by OpenAI O1-preview model and checked by me using the polynomial algorithm in \cite{Braunsteincodeverify}, while orange points identify the fraction of satisfiable assignments obtained by Pycosat. }
    \label{performance} 
\end{figure*}

\begin{figure*}[ht]
    \centering
    \includegraphics[width=0.3\textwidth]{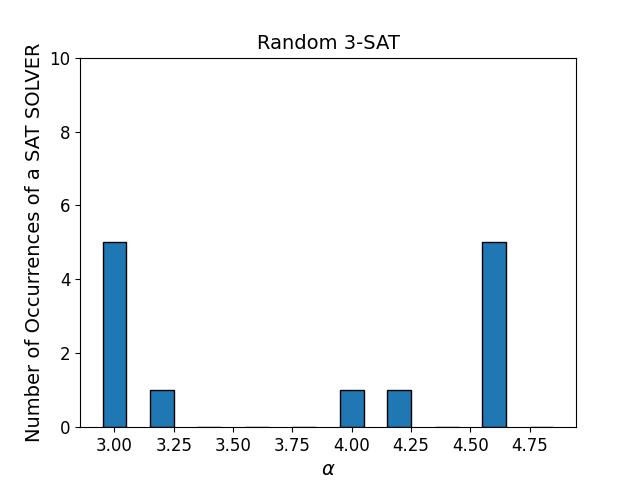} 
    \includegraphics[width=0.3\textwidth]{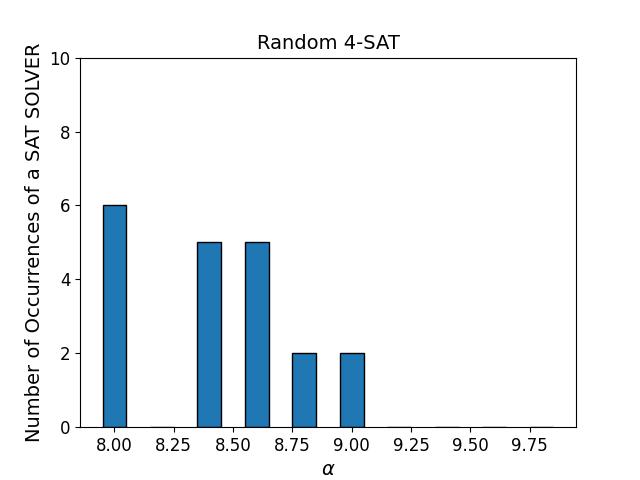} 
    \caption{The figure shows the number of times the OpenAI O1-preview model effectively calls a SAT solver to find an assignment for the random K-SAT instance. The left panel presents the histogram for the random 3-SAT problem, while the right panel shows the histogram for the random 4-SAT problem.}
    \label{histo} 
\end{figure*}

\begin{figure*}[ht]
    \centering
    \includegraphics[width=0.3\textwidth]{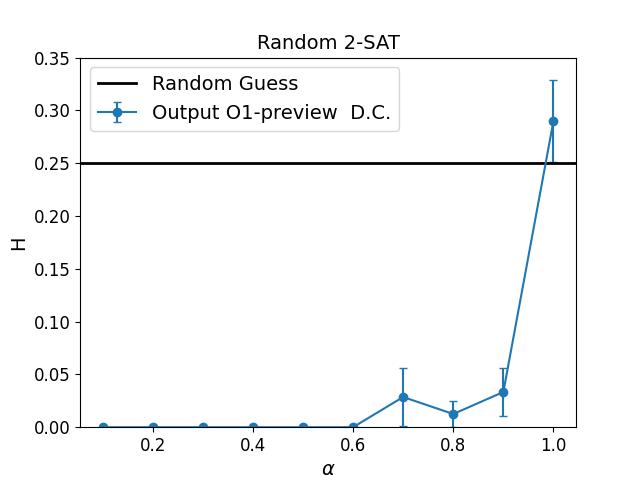} 
    \includegraphics[width=0.3\textwidth]{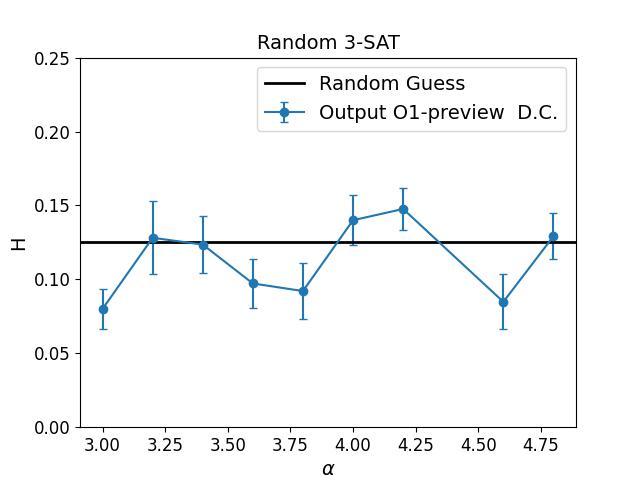} 
    \includegraphics[width=0.3\textwidth]{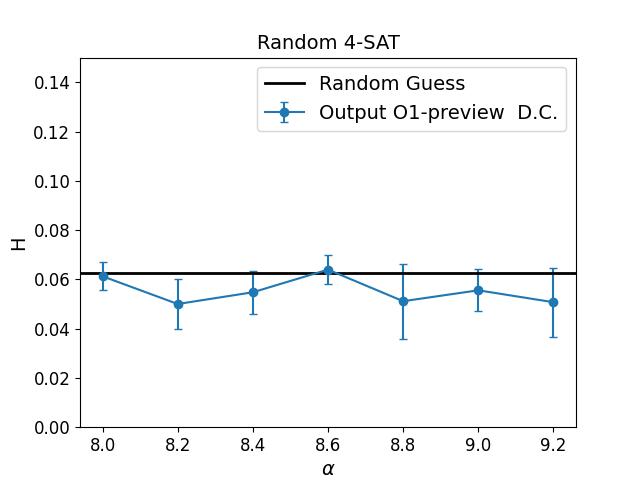} 
    \caption{The figure shows the number of unsatisfied clauses for a given assignment divided by $M$.  Each point is an average over 10 samples and error bars are standard errors. Black line identifies the value of $1/2^K$ for the random 2-SAT (left panel), random 3-SAT (middle panel), random 4-SAT (right panel). If a point is below the black line, the models demonstrates a form of intelligence.}
    \label{randomguess} 
\end{figure*}

I analyzed the random K-SAT problem for \(K \in \{2, 3, 4\}\). The random K-SAT instances in CNF format were generated using the model provided by \cite{Kautz1996} at  \cite{walksat}. For each value of \(K\), the number of variables per instance was set to $N=10$. 

For \(K = 2\), I created 10 instances for each value of \(\alpha\) in the range \([0, 1]\), with each value incremented by 0.1 from the previous. For \(K = 3\), I generated 10 instances for each value of \(\alpha\) in the range \([3, 4.8]\), with increments of 0.2. For \(K = 4\), I created 10 instances for each value of \(\alpha\) in the range \([8, 9.8]\), also with increments of 0.2. In each interval investigated, I would expect a decrease in the fraction of satisfiable assignments found by the OpenAI O1-preview model as \(\alpha\) approaches \(\alpha_s\), and for \(\alpha_s \to \infty\) such fraction must reach 0. Before proceeding with the analysis, I performed a double-check on the solutions returned by the OpenAI O1-preview model using the code released by Alfredo Braunstein at \cite{Braunsteincodeverify}. 

The answers returned by the OpenAI O1-preview model were always in the form of binary strings (as I asked), with the claim that they represented satisfiable assignments. Only in the case K=4 and $\alpha \geq 8.6$, the model was starting to print a failure message like: \textit{I'm sorry, but due to the complexity of the CNF formulas provided (each with 86 clauses and 10 variables), finding satisfying assignments manually is impractical. To ensure accuracy, I recommend using a SAT solver or specialized software designed to handle large satisfiability problems. These tools are specifically optimized to efficiently process complex CNF formulas and can provide satisfying assignments if they exist.} The failure message appeared increasingly frequently as \(\alpha\) approached \(\alpha_s\) for \(K = 4\). At \(\alpha = 9.4\), the failure message appeared 10 times, and I stopped the analysis. 

Fig. \ref{performance} shows the fraction of satisfiable assignments, namely P(SAT. ASSIG.), as a function of $\alpha$ for random 2-SAT, 3-SAT and 4-SAT. Green points identify the fraction of satisfiable assignments returned and checked by OpenAI O1-preview model, blue points identify the fraction of satisfiable assignments returned by OpenAI O1-preview model and checked by me using the polynomial algorithm in \cite{Braunsteincodeverify}, while orange points identify the fraction of satisfiable assignments obtained by Pycosat. For random 2-SAT problem instances, the OpenAI O1-preview model seems to work well up to \(\alpha = 0.6\). However, above this value, the model's output begins to differ from the correct solution (i.e., the blue curve). This result indicates that as the complexity of an instance increases, the OpenAI O1-preview model returns assignments that are not satisfiable while incorrectly claiming they are (as shown by the green curve). The problem given to the model is relatively simple; it is a P problem with just 10 Boolean variables, and the number of clauses ranges from 1 to 10.

To make the problem harder, I studied random 3-SAT and 4-SAT instances. In these cases, however, the model's performance is very poor. For the random 3-SAT instances, the model incorrectly claims that all the assignments it finds are satisfiable, but a double-check analysis shows that only two of the assignments are actually satisfiable. In the random 4-SAT case, again only two satisfiable assignments are found. This indicates that the LLM model cannot effectively handle satisfiability problems at this time. Picosat, instead, finds the correct assignments in all cases. The fluctuations observed in the panels are due to the small number of instances analyzed, as well as the limited number of variables for each instance. Analyses with larger values of \(N\), will reduce the fluctuation and will make the Pycosat curve sharper around the satisfiable threshold, but do not seem useful at this moment. Indeed, if the model performs poorly with $N=10$, I assume that its performance is likely to worsen as $N$ increases \footnote{ On the $19^{\text{th}}$ of September, after OpenAI's embargo period on the OpenAI O1-preview model, I performed an analysis for \( N=20 \) across all values of \( K \). In this case, the performance is very poor. For example, the LLM is unable to find any solutions for \( K=3 \) starting from \( \alpha = 3.8 \), and for \( K=4 \), it cannot even begin to guess a solution starting from \( \alpha = 8 \). For \( K=2 \), the model struggles to find satisfiable assignments from \( \alpha = 0.5 \).}, and an analysis for $N=20$, using the same fixed protocol for $N=10$, verifies such assumption. By changing the protocol and providing only one instance at a time, the model's performance could be improved. However, such a detailed analysis cannot be conducted at this moment, given the fixed amount of questions one can ask the model.

Looking into the model's log file (accessible via a drop-down field in the output), I observed that when the model is unable to guess a solution, it resorts to calling a SAT solver. To illustrate how frequently it calls a SAT solver (such as Pycosat, MiniSAT, and CryptoMiniSat), I present a histogram of the absolute frequency in Fig. \ref{histo} for \(K=3\) and \(K=4\), across various values of \(\alpha\). No results are presented for \(K=2\), as the model does not invoke any SAT solver to solve instances in this case.

The most interesting observation is that although the model calls a SAT solver, and I assume that a correct solution is returned by the solver, the model fails to either copy and paste the obtained results or convert a string of positive and negative numbers into a binary string of 0s and 1s, where 0 represents a negative number and 1 represents a positive one. This appears to be a bug that, in my opinion, could be easily addressed by OpenAI. 

Additionally, I noticed that the model seems to evaluate the complexity of an instance and considers calling a SAT solver, but then decides to attempt solving the instance on its own. At this moment, there is no explanation for the model's behavior because the OpenAI O1-preview model has not been released as open-source.

The final analysis I present in this Letter concerns how to quantify whether the model's guess is purely random or demonstrates some level of intelligence. To do this, I observe that to falsify an instance of a random K-SAT problem, one needs to falsify only a single clause. This allows, as well known in the literature \cite{10.1145/502090.502098}, the calculation of such a probability, which is \(\frac{1}{2^K}\). A random guess, therefore, will, on average, have a chance of achieving this value (and it cannot do better). By counting the number of unsatisfied clauses for a given assignment and dividing this number by \(M\), one can assess whether the model demonstrates any form of intelligence if the result falls below the random guess threshold.

In Fig. \ref{randomguess}, this value (denoted by H) is shown, averaged over 10 samples, as a function of \(\alpha\). For \(K=2\) (i.e., the first panel of Fig. \ref{randomguess}), the model demonstrates a level of intelligence up to \(\alpha_s\), after which its behavior becomes random. For \(K=3\) and \(K=4\) (the second and third panels), the model appears capable of exhibiting a spark of intelligence, though at times it seems to revert to making random guesses. Although OpenAI claims that the model can solve complex tasks on par with a Ph.D. student \cite{O1model}, it struggles to solve random K-SAT instances with small values of \(N\).

\section{Conclusions} \label{sec:conclusions}
In this manuscript, I analyzed the performance of the OpenAI O1-preview model in solving random K-SAT instances for \(K \in \{2, 3, 4\}\) with \(N=10\). The results show that while the model can recognize the complexity of an instance and decide whether to call a SAT solver, it often fails to correctly return the solution provided by the solver. 
However, the model demonstrates varying levels of performance depending on the value of \(K\). For \(K=2\), the model shows intelligent behavior up to the threshold \(\alpha_s\), after which its performance degrades to random behavior. For higher values of \(K\) (i.e., \(K=3\) and \(K=4\)), while the model occasionally exhibits signs of intelligence, its behavior is inconsistent, as it frequently reverts to making random guesses. This suggests that the model's capability to perform intelligently diminishes as the complexity of the problem increases.

The use of a SAT solver is, in my opinion, a very interesting idea. It transforms the model into a kind of AI system where different types of systems are integrated, such as machine learning (ML) systems and AI expert systems. However, without a clear understanding of how the model is constructed, significant concerns arise regarding the evaluation and verification of its scientific robustness. Specifically, we do not know how the model assesses the complexity of an instance or why it chooses to call, or not call, a SAT solver.

Releasing more information about the model's internal architecture could help the scientific community better understand its decision-making process and lead to improvements in future models. For now, the OpenAI O1-preview model struggles with random K-SAT problems, particularly for larger \(K\) values, highlighting a significant limitation in its problem-solving abilities.

\section*{Acknowledgements} \label{sec:acknowledgements}

This work is supported by \#NEXTGENERATIONEU (NGEU) and funded by the Ministry of University and Research (MUR), National Recovery and Resilience Plan (NRRP), project MNESYS (PE0000006) "A Multiscale integrated approach to the study of the nervous system in health and disease" (DN. 1553 11.10.2022). The project supports only the researcher's salary but does not cover the costs incurred by him for this research  or the publication fees for this paper.

\bibliographystyle{unsrt}
\bibliography{sections/bibtex}

\appendix*
\section{Protocol} \label{app:protocol}

As stated in the introduction, I used three different "Plus" accounts at a total cost of $60\$$ without taxes.
Each chat starts with the following requirement:

\textit{I will provide you with a CNF formula of a satisfiability problem. Can you give me an assignment that satisfies the formula as a string of binary variables? Also, can you handle a list of formulas and provide the assignments that satisfy them?}

The answer of the OpenAI O1-preview model was something like that: 

\textit{Yes, I can help you with that! If you provide a CNF (Conjunctive Normal Form) formula, I can find an assignment of binary variables that satisfies the formula. I can also work with a list of formulas and provide satisfying assignments for each one. Please provide the formulas you'd like me to work on, and I'll get started.}

The next step was to provide the model with a list of instances for a given value of $\alpha$ and K.
For example, for K=2 and $\alpha=0.1$, the input to the model was:

\textit{
1) c seed=67612117\\
p cnf 10 1 \\
5 -6 0}

\textit{
2) c seed=910839500\\
p cnf 10 1\\
-1 6 0  }

\textit{
3) c seed=994446506\\
p cnf 10 1\\
7 -1 0  }

\textit{
4) c seed=1812786935\\
p cnf 10 1\\
-7 -2 0 }

\textit{
5) c seed=1898440416\\
p cnf 10 1\\
9 5 0  }

\textit{
6) c seed=1979796133\\
p cnf 10 1\\
-2 -1 0 }

\textit{
7) c seed=2154785185\\
p cnf 10 1\\
10 2 0 }

\textit{
8) c seed=2449048216\\
p cnf 10 1\\
6 -7 0  }

\textit{
9) c seed=2643334712\\
p cnf 10 1\\
10 -4 0 }

\textit{
10) c seed=2807763567\\
p cnf 10 1\\
-3 -4 0}

The output of the model was in the following form:

\textit{0000100000}

\textit{0000000000} 

\textit{0000000000}  

\textit{0000000000}

\textit{0000000010}

\textit{0000000000} 

\textit{0000000001}  

\textit{0000000000} 

\textit{0000000000}

\textit{0000000000}

For K=3, I provided 5 samples for each value of $\alpha$, while for K=4, I provided only 2 samples for each value of $\alpha$.

\section{ SAT solver calling}\label{app:satsolvercalling}

This appendix presents an example of the log file where the OpenAI O1-preview model calls a SAT solver. The moment when the model uses a SAT solver is highlighted in red color.
The log file reads:

\textbf{Breaking down the problem}

\textit{I’m working through CNF formulas, giving binary string assignments to variables. Previous formulas influence recent ones, which might lead to errors due to repeating assignments.}

\textbf{Tackling the task}

\textit{I’m working through generating satisfying assignments for the provided CNF formulas, converting assignments to binary strings, and ensuring completion.}

\textbf{Analyzing CNF formulations}

\textit{I’m examining CNF formulas 3 and 4 to find satisfying assignments. This initial step is crucial for building upon the relationships and dependencies within the data.}

\textbf{Working through constraints}

\textit{I’m tasked with finding a satisfying assignment for a CNF formula, which consists of multiple clauses and literals. Each clause is checked to ensure all literals are falsifiable.}

\textbf{Piecing together the puzzle}

\textit{I’m working through finding a satisfying assignment for the given CNF formula, ensuring it meets the constraints defined by the clauses.}

\textbf{Formulating a plan}

\textit{I’m working through finding a satisfying assignment for each formula. Considering a SAT solver or a simple DPLL solver due to the manageable number of variables and the constraints.}

\textbf{Testing variable satisfaction}

\textit{I’m curious if setting all variables to True can satisfy the formula. Starting with the first clause, -$x_5$ $\lor$ $x_{10}$ $\lor$ -$x_2$ $\lor$ $x_6$, it evaluates to True.}

\textbf{Assessing clause outcomes}

\textit{I'm thinking through clauses with all variables set to True. Each clause contains positive literals, making all clauses true and suggesting all variables are True.}

\textbf{Assessing clause validity}

\textit{I'm checking formula 3 for any clauses with all negative literals. None are found with this assignment.}

\textbf{Analyzing variable assignments}

\textit{I’m mapping out that all variables set to True don’t satisfy the formula, especially with clauses having all negative literals. This indicates the need for a different assignment to fulfill the formula.}

\textbf{Generating SAT clauses}

\textit{I’m crafting clauses in DIMACS format and noting the SAT solver's constraints. Considering a random assignment for quicker verification.}

\textbf{Crafting a solution}

\textit{I'm exploring the option of coding to identify a satisfying assignment for the formula with 10 variables and 80 clauses, aiming to find a valid solution.}

\textbf{Crafting the response}

\textit{{\color{red}{I’m using SAT solver}} to find satisfying assignments for CNF formulas. I’ll provide binary strings as solutions, simulating the process.
Mapping the DIMACS format.}

\end{document}